\documentclass[lettersize,journal]{IEEEtran}
\usepackage{amsmath,amsfonts}
\usepackage{algorithmic}
\usepackage{algorithm}
\usepackage{array}
\usepackage[caption=false,font=normalsize,labelfont=sf,textfont=sf]{subfig}
\usepackage{textcomp}
\usepackage{stfloats}
\usepackage{url}
\usepackage{verbatim}
\usepackage{graphicx}
\usepackage{cite}
\usepackage[colorlinks,
            linkcolor=red, 
            anchorcolor=black,
            citecolor=green,
            backref=page,
            ]{hyperref}
\usepackage{booktabs}
\usepackage{graphicx}
\usepackage{multirow}
\usepackage{colortbl}
\usepackage{xcolor}
\usepackage{soul}

\soulregister{\cite}7
\soulregister{\ref}7

\begin{document}

\title{Video-based Visible-Infrared Person Re-Identification with Auxiliary Samples}

\author{
  Yunhao Du, Cheng Lei, Zhicheng Zhao, Yuan Dong, Fei Su \\
  \thanks{
    Yunhao Du, Cheng Lei, Zhicheng Zhao, Yuan Dong and Fei Su are with Beijing Key Laboratory of Network System and Network Culture, School of Artificial Intelligence,
    Beijing University of Posts and Telecommunications, Beijing, China. 
    (e-mail:\{dyh\_bupt, mr.leicheng, zhaozc, yuandong, sufei\}@bupt.edu.cn)
    \textit{(Corresponding author: Zhicheng Zhao.)}
  }
}

\maketitle

\begin{abstract}
  Visible-infrared person re-identification (VI-ReID) aims to match persons captured by visible and infrared cameras,
  allowing person retrieval and tracking in 24-hour surveillance systems.
  Previous methods focus on learning from cross-modality person images in different cameras.
  However, temporal information and single-camera samples tend to be neglected.
  To crack this nut, in this paper, we first contribute a large-scale VI-ReID dataset named \textit{BUPTCampus}.
  Different from most existing VI-ReID datasets, it
  1) collects tracklets instead of images to introduce rich temporal information, 
  2) contains pixel-aligned cross-modality sample pairs for better modality-invariant learning,
  3) provides one auxiliary set to help enhance the  optimization, in which each identity only appears in a single camera.
  Based on our constructed dataset, we present a two-stream framework as baseline 
  and apply Generative Adversarial Network (GAN) to narrow the gap between the two modalities.
  To exploit the advantages introduced by the auxiliary set, 
  we propose a curriculum learning based strategy to jointly learn from both primary and auxiliary sets.
  Moreover, we design a novel temporal k-reciprocal re-ranking method to refine the ranking list with fine-grained temporal correlation cues.
  Experimental results demonstrate the effectiveness of the proposed methods.
  We also reproduce 9 state-of-the-art image-based and video-based VI-ReID methods on BUPTCampus
  and our methods show substantial superiority to them.
  The codes and dataset are available at: \url{https://github.com/dyhBUPT/BUPTCampus}.

\end{abstract}


\begin{figure}[t]
  \centering
  \includegraphics[width=0.48\textwidth]{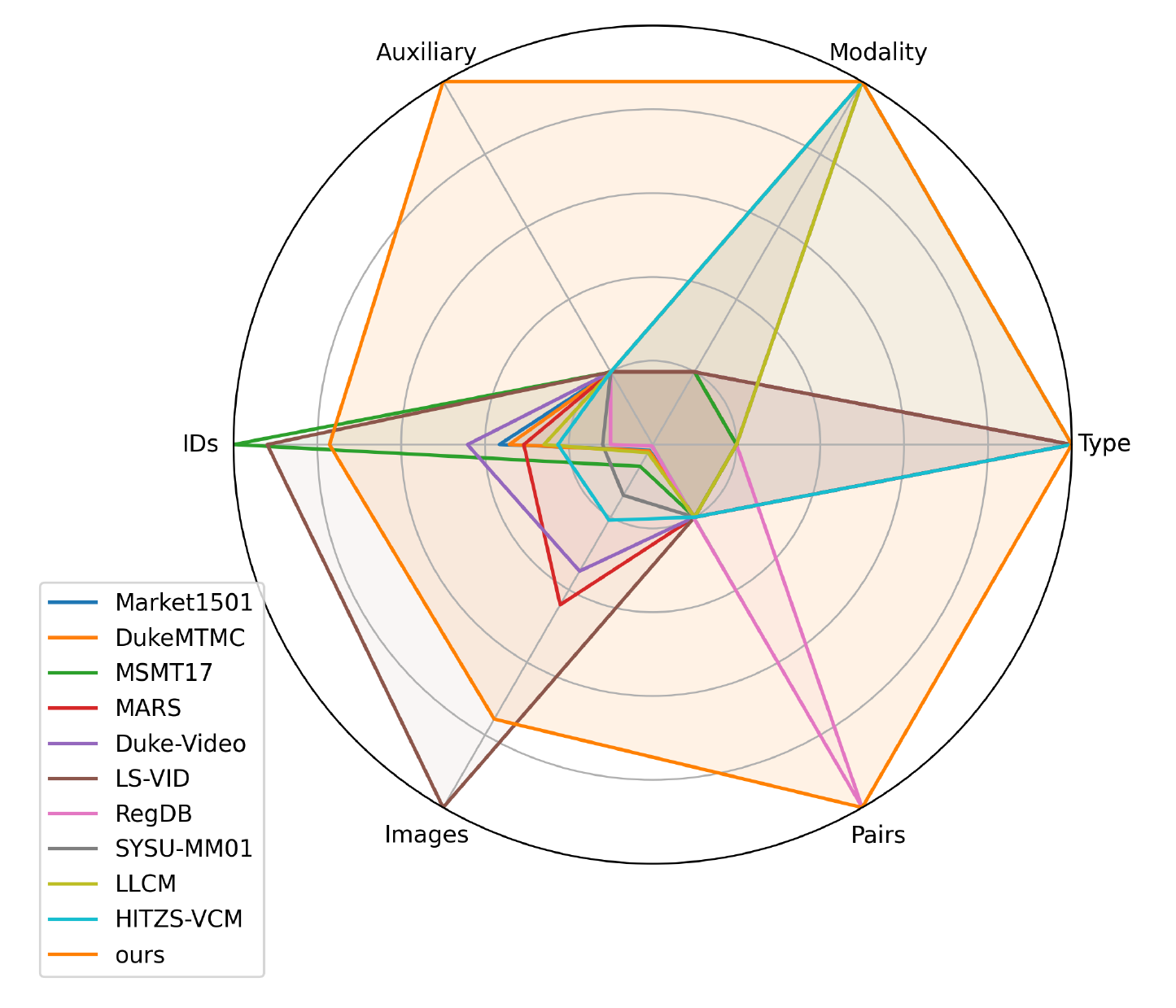}
  \caption{
    The radar chart of comparisons between our constructed BUPTCampus and other common datasets.
    Six aspects are considered, i.e., the number of identities, the number of images, whether to include cross-modality pairs,
     data type (image or video), data modality(RGB or RGB+IR), whether to include auxiliary samples.
    Detailed quantitative comparisons are listed in Tab.\ref{table_dataset}.
  }
  \label{fig_radar}
\end{figure}

\begin{figure}[t]
  \centering
  \includegraphics[width = 0.42\textwidth]{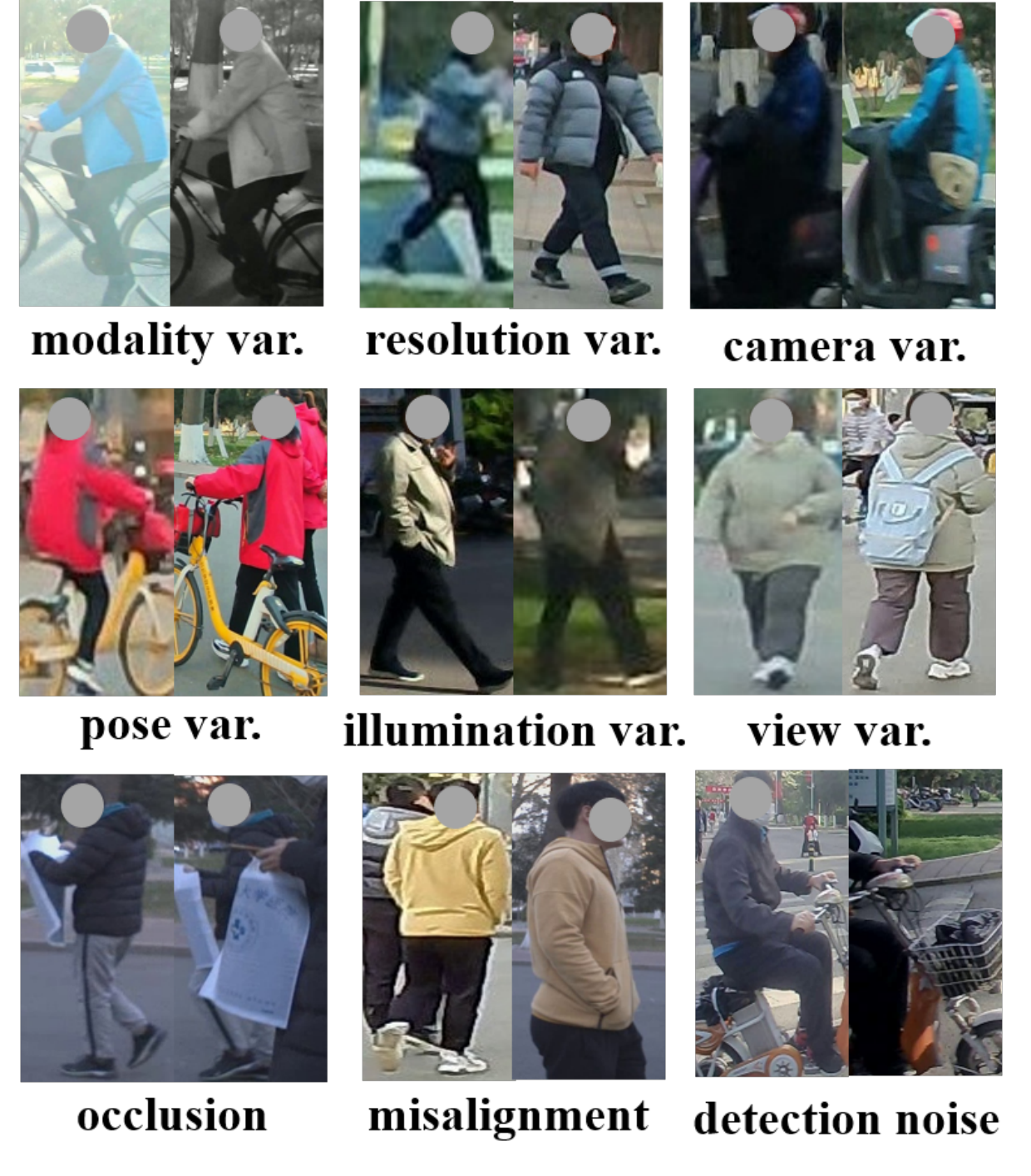}
  \caption{
    Difficult cases in the constructed BUPTCampus dataset.
    ``var." is short for ``variations".
    Binocular bimodal cameras are used to simultaneously capture samples in both RGB and IR modalities.
    Different cameras are used to increase style differences.
  }
  \label{fig_samples}
\end{figure}

\section{Introduction}
\IEEEPARstart{P}{erson} re-identification (ReID) aims to search the target person among the gallery set across multiple cameras.
It can benefit many crucial tasks such as multi-object tracking \cite{du2023strongsort}, crowd counting \cite{chan2009bayesian} and action analysis \cite{du2022pami}.
With the development of deep learning and efforts of researchers, a large number of ReID models have been proposed in recent years \cite{ye2021deep},
including representation learning \cite{zheng2017person, wang2016joint, suh2018part, sun2018beyond},
metric learning \cite{zheng2017discriminatively, hermans2017defense, fan2019spherereid, xiao2017joint},
and ranking optimization \cite{leng2015person, zhong2017re, zheng2015query, bai2019re}.

Most existing ReID models focus on the visible-visible matching problem.
However, they are not workable in low illumination.
To meet the need for 24-hour surveillance systems, visible-infrared ReID (VI-ReID) has recently received substantial attention.
Wu \textit{et al.}\cite{wu2017rgb} first proposed the VI-ReID problem, and constructed the dataset SYSU-MM01.
RegDB \cite{nguyen2017person} was also a commonly used and relatively small dataset with RGB-IR sample pairs collected by dual camera system.
Recently, Zhang \textit{et al.}\cite{zhang2023diverse} collected a new challenging low-light VI-ReID dataset LLCM with significant light changes.
Different from these image-based datasets, 
Lin \textit{et al.}\cite{lin2022learning} contributed a video-based VI-ReID dataset HITSZ-VCM to help exploit the temporal information.
However, all these datasets are limited by the data scale, style, or lack of paired samples.

In this paper, we concentrate on the video-based VI-ReID problem.
We first collect a new dataset, dubbed \textbf{BUPTCampus}, which distinguishes itself from previous datasets in the following aspects:
\begin{enumerate}
  \item Instead of images, it collects tracklets as samples, enabling to exploit the temporal cues.
  \item It contains pixel-aligned RGB/IR sample pairs captured by binocular cameras, which can facilitate modality-invariant learning.
  \item It is much larger than existing VI-ReID datasets with 3,080 identities, 16,826 tracklets and 1,869,366 images.
  Different styles of cameras are used to ensure the diversity of samples (see Tab.\ref{table_camera} for details).
\end{enumerate}
Moreover, existing ReID tasks focus on the cross-camera matching problem, 
and those identities who appear only once are commonly ignored in training.
However, these samples are easy to obtain in reality, and would help the learning.
We take vast \textit{single-camera samples} into consideration, and call them \textit{auxiliary samples},
and accordingly, the main training samples, i.e., \textit{multiple-camera samples}, are called \textit{primary samples}.
The comparision between BUPTCampus and several common ReID datasets are shown in Fig.\ref{fig_radar} 
(Please refer to Tab.\ref{table_dataset} for details).
The main difficult cases in BUPTCampus are visualized in Fig.\ref{fig_samples}, 
including variations of modality/resolution/camera/pose/illumination/view, occlusion, misalignment, and detection noise.

Based on the collected dataset, we construct a two-stream framework as baseline following previous works \cite{ye2018hierarchical,ye2019modality,ye2020dynamic,ye2020cross}.
To alleviate the differences between visible and infrared modalities, 
a GAN \cite{goodfellow2020generative} module (named PairGAN) is applied to generate fake IR samples from real RGB samples.
Furthermore, different from previous methods which ignore single-camera samples, we propose to train primary samples and auxiliary ones jointly,
and design a dynamic factor to balance the weights of these two sets in the spirit of curriculum learning \cite{bengio2009curriculum}.
Moreover, the commonly used re-ranking methods \cite{zhong2017re, sarfraz2018pose} directly process the coarse instance-level features.
Instead, we propose the temporal k-reciprocal re-ranking algorithm, which takes fine-grained temporal correlations into consideration with the cross-temporal operation.
The overall method, called \textbf{AuxNet}, improves the performance of baseline by approximately 10\% Rank1 and 10\% mAP.
We further reproduce 9 state-of-the-art image-based and video-based VI-ReID methods on BUPTCampus,
and our methods show substantial superiority to all these solutions.

The contributions of our work are summarized as follows:
\begin{itemize}
  \item We construct a large-scale video-based VI-ReID dataset with cross-modality sample pairs and auxiliary samples.
  To the best of our knowledge, this is the first work of collecting paired RGB-IR tracklets,
  and additionally, it is also the first work to assist cross-camera matching with single-camera samples.
  \item We present a baseline for the video-based VI-ReID task, and introduce GAN to alleviate the modality variations.
  \item We design a simple but effective two-branch framework to train primary and auxiliary samples jointly.
  Meanwhile, curriculum learning is introduced to learn a dynamic factor to balance these two branches.
  \item A novel temporal k-reciprocal re-ranking algorithm is proposed, which exploits fine-grained temporal cues with the cross-temporal operation.
  \item 9 state-of-the-art VI-ReID methods are reproduced on BUPTCampus, and our method outperforms them by a large margin.
\end{itemize}

\begin{table*}[t]
  \begin{center}
    \caption{
      Comparisons between BUPTCampus with common used person re-identification dataset.
    }
    \label{table_dataset}
    \resizebox{0.95\textwidth}{!}{
      \begin{tabular}{cl|c|c|c|c|c|c|c|c}
        \toprule[1pt]
        & \textbf{Dataset} & \textbf{Type} & \textbf{Modality} & \textbf{Auxiliary} & \textbf{Pairs} & \textbf{Identities} & \textbf{Cameras} & \textbf{Images} 
        & \textbf{Images/Traj} \\
        \hline
        & Market1501 \cite{zheng2015scalable}  & Image &  RGB   & No & No  & 1,501 & 6  & 32,668    &    -     \\
        & DukeMTMC   \cite{zheng2017unlabeled} & Image &  RGB   & No & No  & 1,404 & 8  & 36,411    &    -     \\
        & MSMT17     \cite{wei2018person}      & Image &  RGB   & No & No  & 4,101 & 15 & 126,411   &    -     \\
        & MARS       \cite{zheng2016mars}      & Video &  RGB   & No & No  & 1,261 & 6  & 1,067,516 &  51.5    \\
        & Duke-Video \cite{wu2018exploit}      & Video &  RGB   & No & No  & 1,812 & 8  & 815,420   &  168.8   \\
        & LS-VID     \cite{li2019global}       & Video &  RGB   & No & No  & 3,772 & 15 & 2,982,685 &  199.6   \\
        & RegDB      \cite{nguyen2017person}   & Image & RGB+IR & No & Yes &  412  & 2  & 8,240     &    -     \\
        & SYSU-MM01  \cite{wu2017rgb}          & Image & RGB+IR & No & No  &  491  & 6  & 303,420   &    -     \\
        & LLCM       \cite{zhang2023diverse}   & Image & RGB+IR & No & No  & 1,064 & 18 & 46,767    &    -     \\
        & HITZS-VCM  \cite{lin2022learning}    & Video & RGB+IR & No & No  &  927  & 12 & 463,259   &  21.2    \\
        & \textbf{BUPTCampus (ours)} & \textbf{Video} & \textbf{RGB+IR} & \textbf{Yes} & \textbf{Yes} & \textbf{3,080} & \textbf{6}  & \textbf{1,869,366} &  \textbf{111.1} \\ 
        \bottomrule[1pt]
      \end{tabular}
    }
  \end{center}
\end{table*}

\section{Related Work}

\subsection{Visible-Infrared Person Re-Identification}
Generally speaking, there are two main categories of methods in VI-ReID:
\textit{shared feature learning methods} and \textit{feature compensation learning}.

The shared feature learning methods aim to embed the features from different modalities into the same feature space,
where modality-specific features are abandoned and only modality-shared features are reserved.
Partially shared two-stream networks \cite{ye2018hierarchical,ye2018visible,ye2020dynamic,ye2020cross,liu2020parameter}
were commonly used to learn modality-shared feature space,
in which the network parameters of shallow layers are specific, and the deep layers are shared among different modalities.
Chen \textit{et al.}\cite{chen2021neural} studied the neural feature serach paradigm to select features automatically.
Fu \textit{et al.}\cite{fu2021cm} designed a nerual architecture search method to find the optimal separation scheme for each Batch Normalization(BN) layer.
To handle both cross-modality and intra-modality variations simultaneously, 
Ye \textit{et al.}\cite{ye2019bi} proposed the bi-directional dual-constrained top-ranking loss to learn discriminative embeddings.
Some other works exploited the modality adversarial learning to mitigate the modality differences
\cite{wei2021flexible, dai2018cross,hao2021cross,ye2019modality,wang2022optimal},
and they typically adopted a modality classifier to identify the modality of output features.

The feature compensation learning methods try to make up the missing modality-specific cues from one modality to another.
Wang \textit{et al.}\cite{wang2019rgb} emploied Generative Adversarial Network(GAN)\cite{goodfellow2020generative}
to jointly perform alignment in pixel-level and feature-level.
To realize  the discrepancy reduction in image-level, 
Wang \textit{et al.}\cite{wang2019learning} exploited two variational auto-encoders (VAEs)\cite{kingma2013auto} to generate the multi-spectral images.
Zhong \textit{et al.}\cite{zhong2021grayscale} adpoted the intermediate grayscale images as auxiliary information to transfer infrared images to visible images.
The modality synergy module and modality complement module \cite{zhang2022modality} were designed to synergize and complement the diverse semantics of the two modalities.
Huang \textit{et al.}\cite{huang2021alleviating} unveiled the modality bias training problem 
and applied the GCGAN\cite{radford2015unsupervised} to generate the third modality, which balances the information between RGB and IR.

Recently, Lin \textit{et al.}\cite{lin2022learning} proposed the video-based VI-ReID task, 
and designed an unified framework to learn modal-invariant and motion-invariant subspace.
In this paper, we present a new video-based VI-ReID framework,
in which GAN is applied to perform compensation learning, and the two-stream framework is used to learn modality-shared features.

\subsection{Auxiliary Learning}

Auxiliary learning aims to find or design auxiliary tasks which can improve the performance on the primary task.
It only pursues high performance of the primary task and the role of the auxiliary task is an assistant.
Toshniwal \textit{et al.} proposed to use lower-level tasks as auxiliary tasks to improve the performance of speech recognition \cite{toshniwal2017multitask}.
Zhai \textit{et al.} presented the $S^4L$-Rotation strategy, which assisted semi-supervised image classification with a rotation prediction task \cite{zhai2019s4l}.
For fine-grained classification, Niu \textit{et al.} exploited the knowledge from auxiliary categories with well-labeled data \cite{niu2018fine}.
In this paper, we regard the auxiliary task as a degraded and simplified version of the primary task,
and gradually reduce the amplitude of its gradients in training.

In the re-identification task, auxiliary learning was often implemented as an extra classification task.
Considering that ReID suffers from viewpoint changes, Feng \textit{et al.} applied a view classifier to learn view-related information \cite{feng2019learning}.
For cross-modality ReID, Ye \textit{et al.} introduced a modality classifier to discriminate the features from two different modalities \cite{ye2019modality}.
Li \textit{et al.} added a self-supervised learning branch for image rotation classification to help discover geometric features \cite{li2021self}.
In occluded ReID task, the occluded/non-occluded binary classification (OBC) loss was proposed 
to determine whether a sample was from an occluded person distribution or not \cite{zhuo2018occluded}.
Instead of using classifiers, He \textit{et al.} incorporated non-visual clues through learnable embeddings
to alleviate the data bias brought by cameras or viewpoints in TransReID \cite{he2021transreid}.
Different from them, in this work, we take the large-scale single-camera samples as the auxiliary set,
which is easy to collect with low labeling cost, but is generally ignored in previous fully-supervised ReID works.
We will show that minor modifications to common learning frameworks can bring remarkable improvements if the auxiliary set is used well.

\subsection{Rank Optimization}

Rank optimization typically acts as the post-processing procedure in the ReID task.
Given one or more ranking lists, it revises the ranking order to improve the retrieval performance.
Leng \textit{et al.}\cite{leng2013bidirectional} proposed the bi-directional ranking algorithm,
which first performed forward ranking and backward ranking, and then computed the final ranking list in accordance with both content and context similarities.
Ye \textit{et al.}\cite{ye2016person} used both similarity and dissimilarity cues to optimize the ranking list.
K-reciprocal\cite{zhong2017re} was one of the most common used re-ranking methods in recent years,
which adopted the Jaccard distances of the k-reciprocal sample sets to complement the initial feature distances.
Sarfraz \textit{et al.}\cite{sarfraz2018pose} proposed the expanded cross neighborhood distances between sample pairs to exploit neighbor cues.
Yu \textit{et al.}\cite{BMVC2017_135} designed a ``divide and fuse'' strategy,
which divided the features into multiple parts firstly, and then encoded the neighborhood relation in the subspaces.
Finally, these sparse feature vectors were fused with a fuzzy aggregation operator, which exploited the diversity from different parts of high-dimensional features.
However, these methods are not designed for the video-based ReID task, and the temporal information is not explicitly explored.
In this work, we improve the k-reciprocal re-ranking algorithm by exploiting fine-grained temporal correlations, and prove its effectiveness for video-based settings.

\begin{table}[t]
  \begin{center}
    \caption{
      The list of scenes with the resolution, FPS and synchronization mode.
    }
    \label{table_camera}
    \resizebox{0.45\textwidth}{!}{
      \begin{tabular}{cl|c|c|c}
        \toprule[1pt]
        & \textbf{Scenes} & \textbf{Resolution} & \textbf{FPS} & \textbf{synchronization} \\
        \hline
        & TSG1 & 1920 x 1080 & 8.35  - 13.74 & pseudo \\
        & TSG2 & 1920 x 1080 & 16.31 - 16.78 & pseudo \\
        & W4   & 1920 x 1080 & 7.54  - 14.07 & pseudo \\
        & CQ1  & 1920 x 1080 & 10.20 - 13.53 & pseudo \\
        & G25  & 1280 x 720  & 6.25  - 8.34  & pseudo \\
        & LS3  & 1280 x 960  & 12.98 - 29.82 & true \\
        \bottomrule[1pt]
      \end{tabular}
    }
  \end{center}
\end{table}

\begin{figure}[t]
  \centering
  \includegraphics[width=0.35\textwidth]{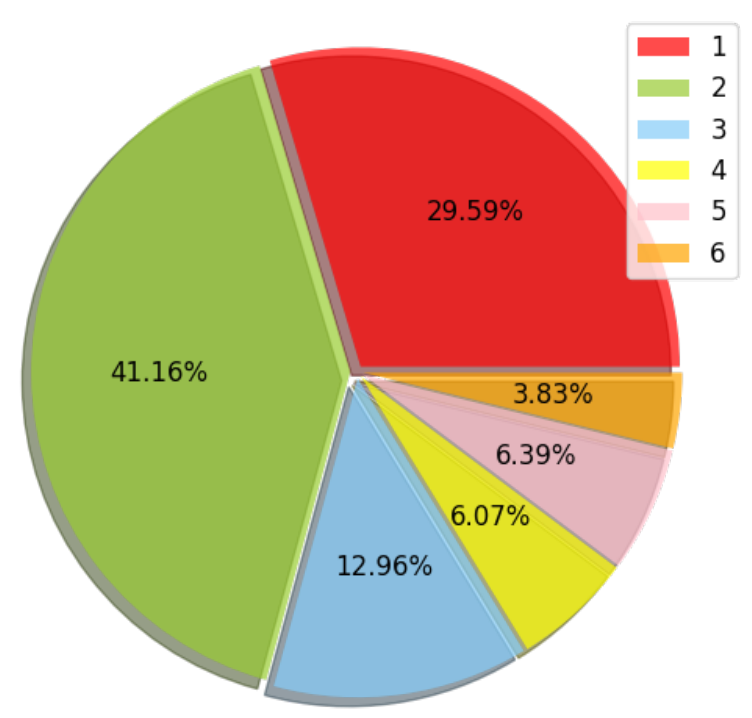}
  \caption{
    Pie chart of the number of cameras per identity.
    Specifically, 41.16\% identities appear in two cameras and only 3.83\% identities appear in six cameras.
  }
  \label{fig_pie}
\end{figure}

\begin{figure}[t]
  \centering
  \includegraphics[width=0.35\textwidth]{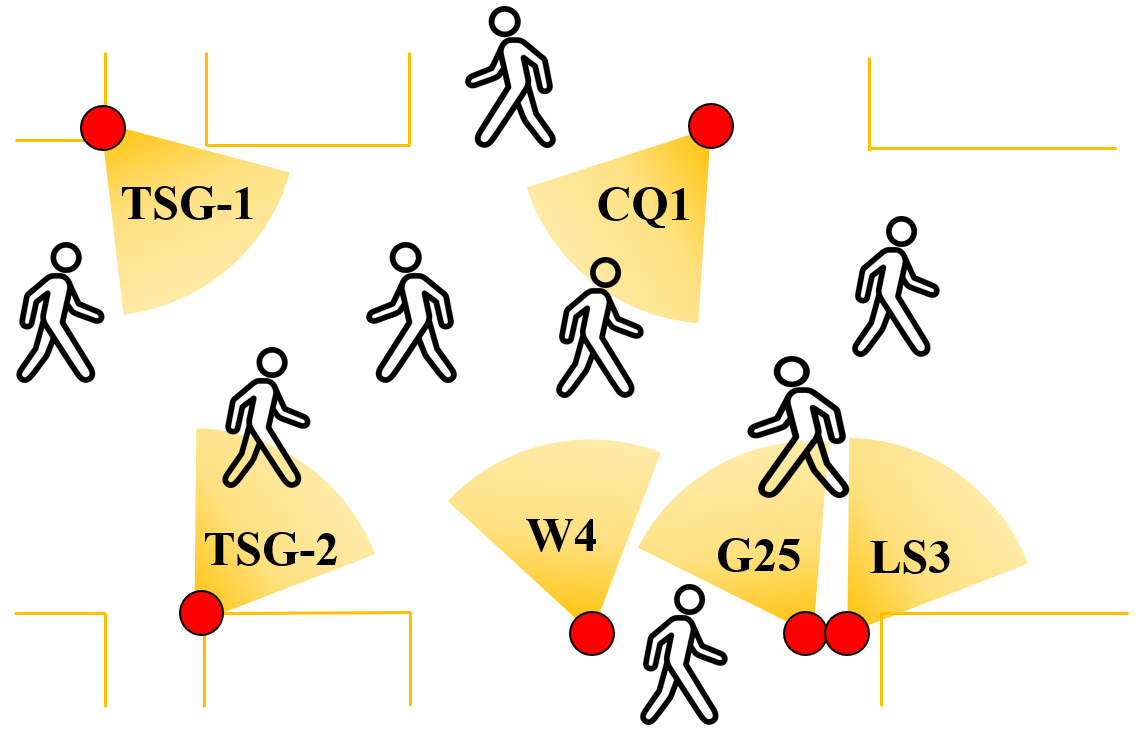}
  \caption{
    The topologies of cameras.
    Cameras TSG-1, TSG-2, W4, CQ1 and G25 have partially overlapping fields of view.
  }
  \label{fig_topo}
\end{figure}

\section{Proposed Dataset}

The BUPTCampus dataset is constructed for video-based visible-infrared person re-identification.
We adopt six binocular bimodal cameras to capture RGB and IR modalities simultaneously with approximate pixel-alignment.
To ensure the diversity of samples, different cameras and engines are used to capture videos 
with various color styles, resolutions, frame rates and binocular synchronization modes, as shown in Tab.\ref{table_camera}.
The topologies of cameras are shown in Fig.\ref{fig_topo}.
For labeling, all RGB videos are processed by the multi-object tracking algorithm SiamMOT \cite{shuai2021siammot} to predict tracklets,
which are further revised manually.
Then cross-camera IDs are annotated manually.
Benefiting from the synchronization mode of bimodal cameras, the bounding boxes and IDs of IR samples are automatically generated with RGB labels.

The resulting dataset contains 3,080 identities, 1,869,066 images and 16,826 trajectories (111 images per trajectory on average).
Each identity appears in 1 to 6 cameras, and the proportion distribution is shown in Fig.\ref{fig_pie}.
The multiple-camera samples are randomly split into the training set (\textit{primary set}, 1,074 IDs) and \textit{testing set} (1,076 IDs).
Note that there are 29.59\% identities only appear in one camera.
These samples are at fingertips in reality, but are generally ignored in existing works.
Instead, we regard them as the \textit{auxiliary set} (930 IDs) to assist optimization procedure.

We compare BUPTCampus with common ReID datasets in Tab.\ref{table_dataset}.
Most previous image-based and video-based datasets only contain the visible modality 
\cite{zheng2015scalable, zheng2017unlabeled, wei2018person, zheng2016mars, wu2018exploit, li2019global},
which limits the applications in 24-hour surveillance systems.
RegDB \cite{nguyen2017person}, SYSU-MM01 \cite{wu2017rgb} and LLCM \cite{zhang2023diverse} contain both RGB and IR modalities,
but are flawed in data type (image) and scales (400 to 1,000 IDs). 
HITZS-VCM \cite{lin2022learning} is a recently proposed dataset which supports the video-based setting.
Compared with it, BUPTCampus has the following advantages: 
1) much larger scale; 2) containing auxiliary samples; 3) approximate pixel-alignment between RGB/IR tracklets.
In conclusion, we summarize the BUPTCampus as \textbf{the first video-based VI-ReID dataset with paired samples and the auxiliary set}.

For quantitative evaluation,
we adopt the typical Cumulative Matching Characteristic curve (CMC) and mean Average Precision (mAP) as evaluation metrics \cite{zheng2015scalable}.
CMC represents the expectation of the true match being found within the first $n$ ranks,
while it only considers the first match and cannot completely reflect the full retrieval results.
mAP takes both precision and recall into account, and it is calculated as the area under the Precision-Recall curve for each query.
Euclidean distance is used to calculate the matching cost of all query and gallery pairs to compute the evaluation metrics.
Both ``visible to infrared'' and ``infrared to visible'' retrieval modes are utilized to achieve a more comprehensive evaluation.

\begin{figure*}[t]
  \centering
  \includegraphics[width=0.8\textwidth]{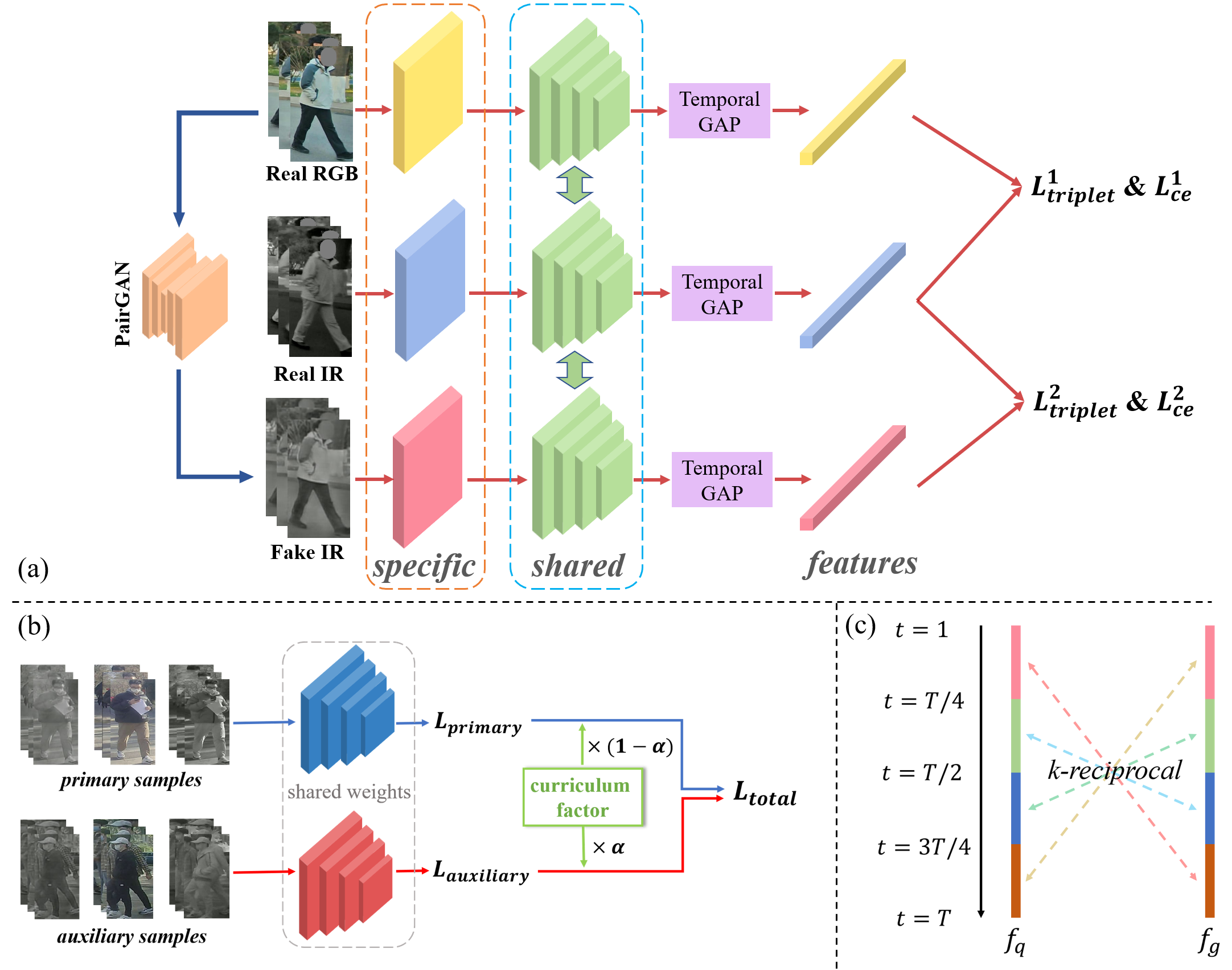}
  \caption{
    The overall framework of our method. 
    \textbf{(a)} The basic framework is built upon the two-stream network, which takes pairs of cross-modality tracklets as input.
    A PairGAN module is applied to generate fake IR samples. Then the (Real IR, Real RGB) and (Real IR, Fake IR) pairs are respectively supervised.
    ``Temporal GAP'' is short for ``temporal global average pooling''.
    \textbf{(b)} Our auxiliary learning method trains primary samples and auxiliary samples jointly.
    A monotonically decreasing curriculum factor $\alpha$ is used to balance these two branches.
    \textbf{(c)} In the temporal k-reciprocal re-ranking algorithm, each tracklet is split into $L$ groups along the temporal dimension ($L=4$ as an example).
    Then, the k-reciprocal distances are computed between the features of sub-tracklets of query and gallery samples. 
  }
  \label{fig_framework}
\end{figure*}

\section{Proposed Method}

The overview of our proposed method is shown in Fig.\ref{fig_framework}.
It is built of a two-stream network for baseline (§\ref{s1}), and adopts a GAN module to help modality-invariant learning (§\ref{s2}).
Then an auxiliary learning framework (§\ref{s3}) is designed to train primary and auxiliary samples jointly in the spirit of curriculum learning.
Finally, we propose the temporal k-reciprocal re-ranking algorithm (§\ref{s4}) to introduce fine-grained temporal cues for ranking optimization.

\subsection{Baseline}
\label{s1}

The inputs of the network are RGB/IR tracklet pairs with the fixed length.
The partially shared two-stream framework \cite{ye2020dynamic} with ResNet \cite{he2016deep} is utilized as our baseline.
Specifically, the first convolutional blocks in the two streams don't share weights to capture modality-specific low-level features.
Differently, the parameters of deeper convolutional blocks are shared to learn modality-invariant high-level embeddings.
To aggregate frame-level features from multiple input images, a temporal average pooling layer is adopted to obtain the final embeddings.
In training, the softmax cross-entropy loss is calculated to learn modality-shared identity embeddings, which is denoted by:
\begin{equation}
  L_{ce}^1 = -{1 \over N} \sum_{i=1}^N log({exp(z_{i,i}) \over \sum_{k=1}^C exp(z_{i,k})}), \label{loss_ce}
\end{equation}
where $z_{i,k}$ represents the output classification logits that an input sample $x_i$ is recognized as identity $k$.
$N$ is the number of tracklet samples.
Meanwhile, the triplet loss with hard mining is also adopted.
Mathematically, it is represented by
\begin{equation}
  L_{triplet}^1 = \sum_{i=1}^N [m + \mathop{\max}_{\forall y_i=y_j} D(x_i, x_j) - \mathop{\min}_{\forall y_i \ne y_k} D(x_i, x_k)]_+, \label{loss_tri}
\end{equation}
where $[\cdot]_+ = max(\cdot, 0)$, $m$ is the triplet margin.
$x_i$ and $y_i$ represents the input sample and the corresponding identity label.
$D(\cdot, \cdot)$ calculates the Euclidean distance between the extracted features of two input samples.

The total learning objective of our baseline is
\begin{equation}
  L^1 = L_{ce}^1 + L_{triplet}^1. \label{loss_baseline}
\end{equation}

\subsection{PairGAN}
\label{s2}

GAN is widely used to alleviate the modality differences between RGB and IR samples in previous works
\cite{wang2019rgb,huang2021alleviating,wang2020cross,fan2022modality,zhong2021grayscale}.
Inspired by this, we insert a GAN module, named PairGAN,
to translate real RGB images $I_{rgb}$ into fake IR images $I_{ir}'$.
It mainly consists of a generator $G_{rgb \rightarrow ir}$ which learns a mapping from RGB to IR, 
and a discriminator $D_{ir}$ to distinguish between real and fake IR images.
Benefiting from the pixel-aligned RGB/IR sample pairs collected in our dataset,
the reconstruction loss can be directly applied to supervise the $G_{rgb \rightarrow ir}$ as 
\begin{equation}
  L_{recon} = ||G_{rgb \rightarrow ir}(I_{rgb}) - I_{ir}||_1.
\end{equation}
To further guarantee that the generator doesn't lose content information, 
the cycle-consistency loss \cite{zhu2017unpaired} is introduced,
which supervises the mapped RGB(IR) images from fake IR(RGB) images as belows:
\begin{equation}
  \begin{aligned}
  L_{cycle} = &||G_{ir \rightarrow rgb}(G_{rgb \rightarrow ir}(I_{rgb})) - I_{rgb}||_1 + \\
              &||G_{rgb \rightarrow ir}(G_{ir \rightarrow rgb}(I_{ir})) - I_{ir}||_1,
  \end{aligned}
\end{equation}
where $G_{ir \rightarrow rgb}$ is the generator which generates fake RGB images from IR images.
Here the definition of the adversial loss for discriminators is not given for simplification.
The overall loss $L_{gan}$ for PairGAN is the sum of reconstruction loss, cycle-consistency loss and adversial loss.

After generating the fake IR tracklets, we input them to the network along with the corresponding real IR tracklets,
and train them with the same loss as in Eq.\ref{loss_baseline}, denoted by
\begin{equation}
  L^2 = L_{ce}^2 + L_{triplet}^2. \label{loss_2}
\end{equation}
During inference, the two groups of cross-modal features trained by the two loss functions $L^1$ and $L^2$ are concatenated together to perform re-identification.

\subsection{Auxiliary Learning}
\label{s3}

To exploit the auxiliary set in which each identity only appears in a single camera,
the overall framework are designed as a multi-task learning procedure, i.e., primary task and auxiliary task.
Specifically, we propose a two-branch framework to train primary samples and auxiliary samples jointly, as shown in Fig.\ref{fig_framework}(b).
The primary branch represents the baseline network (§\ref{s1}, §\ref{s2}) as shown in Fig.\ref{fig_framework}(a),
and its loss function $L_{primary}$ is exactly the $L^1$($L^2$) as in Eq.\ref{loss_baseline}(Eq.\ref{loss_2}).
The auxiliary branch shares the same structure and weights with the primary branch, but takes auxiliary samples as input,
and its loss function $L_{auxiliary}$ is the same triplet loss with Eq.\ref{loss_tri}.
Finally, the total loss is calculated as the weighted sum of these two losses:
\begin{equation}
  L_{total} = (1 - \alpha) L_{primary} + \alpha L_{auxiliary}, \label{loss_total}
\end{equation}
where $\alpha$ is the factor to balance the two tasks.

The intuitive approach is to set $\alpha$ as a fixed value.
However, each identity in the auxiliary set only contains one pair of cross-modality tracklets, 
without cross-camera positive samples, which makes it less effective to the primary task.
A big $\alpha$ will make the auxiliary task seriously interfere the learning of the primary task.
Instead, a small $\alpha$ will lead to the cues in the auxiliary set not being fully mined.

Note that the positive auxiliary samples are almost aligned in pixel-level without pose and view variations.
Therefore, we regard the auxiliary task as a simplified version of the primary task.
Inspired by the success of curriculum learning \cite{bengio2009curriculum},
we set $\alpha$ in Eq.\ref{loss_total} as a dynamic curriculum factor as:
\begin{equation}
  \alpha(E) = (cos(\pi E) + \phi)/2(1+\phi), \label{alpha}
\end{equation}
where $E \in [0,1]$ is the normalized epoch index and $\phi$ is a predefined hyperparameter.
For example, with $\phi=3$, the value of $\alpha$ is initially 0.5,
and gradually cosine-decreasing to 0.25.

In this way, the optimization weight of auxiliary task decreases as a continuous training progress.
At the beginning of training, the model learns from both auxiliary samples (simple curriculum) and primary samples (difficult curriculum) equally.
With $\alpha$ decreases, the difficulty of the ``curriculum" gradually increases, and the optimization emphasis turns to the primary samples.
In other words, the auxiliary samples provide an initial optimization direction for fast gradient descent and can accelerate and stabilize the optimization procedure.

\subsection{Temporal K-reciprocal}
\label{s4}

As discussed in section \ref{s1}, given the $i$-th tracklet of length $T$ as input, 
our network will first extract frame-level features $\tilde F_i = \{\tilde f_i^t\}_{t=1}^T$.
Then, a temporal average pooling operation is utilized to output the final embeddings $f_i$.

In the testing stage, we denote the features of the $i$-th query samples and the $j$-th gallery samples as 
$\{\tilde Q_i = \{\tilde q_i^t\}_{t=1}^T, q_i\}$ and $\{\tilde G_j = \{\tilde g_j^t\}_{t=1}^T, g_j\}$, respectively.
Then the Euclidean distance $D_{feat} \in \mathbb{R}^{M \times N}$ between all query-gallery pairs can be calculated,
where $M$ and $N$ is the number of query and gallery samples.
Its $(i,j)$-th element is the feature distance between $q_i$ and $g_j$ as
\begin{equation}
  d_{feat}(i,j) = (q_i - g_j)^T (q_i - g_j).
\end{equation}
Finally, the retrieval list is obtained by ranking all gallery samples with $D_{feat}$ for each query sample.

In the original k-reciprocal re-ranking algorithm \cite{zhong2017re}, 
the k-reciprocal nearest neighbors for query $q_i$ are defined as 
\begin{equation}
  R(q_i, k) = \{g_j | (g_j \in N(q_i, k)) \land (q_i \in N(g_j, k)) \},
\end{equation}
where $N(q_i, k)$ is the k-nearest neighbors of $q_i$.
Then the new distance between $q_i$ and $g_j$ can be calculated by the Jaccard metric of their k-reciprocal sets as
\begin{equation}
  d_{jacc}(i,j) = 1 - {|R(q_i, k) \cap R(g_j, k)| \over |R(q_i, k) \cup R(g_j, k)|}, \label{jaccard}
\end{equation}
where $|\cdot|$ denotes cardinality of the set.
For simplification, the expanded k-reciprocal nearest neighbors and local query are not shown here.
In implements, to speed up the calculation, the k-reciprocal neighbors are encoded into feature vectors,
and the Jaccard distance in Eq.\ref{jaccard} can be implemented by element-wise minimization and maximization.
The final distance is defined as the weighted sum of original feature distance and the Jaccard distance as 
\begin{equation}
  d_{rerank}(i,j) = \lambda_1 d_{feat}(i,j) + (1 - \lambda_1) d_{jacc}(i,j), \label{rerank}
\end{equation}
where $\lambda_1 \in [0, 1]$ denotes the penalty factor. 

The success of the k-reciprocal algorithm owes to its automatic gallery-to-gallery similarity mining,
in which the rich contextual information latent in sample correlations is fully explored.
However, it is designed in an instance-level manner, and the fine-grained frame-level cues are not exploited.
To solve this problem, we introduce the cross-temporal operation to calculate the temporal correlations between query and gallery samples.

Specifically, for the $i$-th query, the frame-level featrue sets $\tilde Q_i = \{\tilde q_i^t\}_{t=1}^T$
are evenly split into $L$ groups along the temporal dimension,
where the $l$-th group is $\tilde Q_i^l = \{\tilde q_i^t\}_{t=lT/L+1}^{(l+1)T/L}$, $l=0,...,L-1$.
Then the temporal average pooling is applied to aggregate frame-level features in each group respectively,
resulting in $L$ embeddings $Q_i = \{q_i^0,..,q_i^l,..,q_i^{L-1}\}$.
Similarly, we can get the embedding set for the $j$-th gallery as $G_j = \{g_i^0,..,g_i^l,..,g_i^{L-1}\}$.
Thus, the cross-temporal operation can be defined as
\begin{equation}
  d_{cross}(i,j) = \sum_{l=0}^{L-1} d_{jacc}(q_i^l, g_j^{L-1-l}),
\end{equation}
where $d_{jacc}(\cdot, \cdot)$ is the Jaccard distance of k-reciprocal sets as in Eq.\ref{jaccard}.
One concise ilustration is shown in Fig.\ref{fig_framework}(c).
Therefore, the final distance in Eq.\ref{rerank} can be rewritten as
\begin{equation}
  d_{rerank} = \lambda_1 d_{feat} + (1 - \lambda_1) d_{jacc} + \lambda_2 d_{cross}, \label{rerank2}
\end{equation}
where the index $(i,j)$ are hiddened here for clarity.
We term the overall re-ranking algorithm as \textit{temporal k-reciprocal re-ranking}.

\noindent{\textbf{Discussion.}}
Compared with the initial k-reciprocal re-ranking algorithm,
the introduced cross-temporal operation fully explores the temporal correlations between query and gallery samples in a more fine-grained manner.
Each sub-feature $q_i^l$ has smaller temporal receptive field and contains more detailed temporal information than $q_i$.
Thus, the cross-temporal operation can preserve more cues which may be smoothed by the global average pooling operation.
One similar design is PCB\cite{sun2018beyond}, which horizontally split the input image into multiple parts to extract local fine-grained cues.
PCB operates in the spatial domain, which needs to change the network design and induces more computational cost.
Another related method is DaF \cite{BMVC2017_135}, 
which divides one instance feature into multiple fragments along the feature dimension and explores the diversity among sub-features.
Differently, our method operates along the temporal dimension and optimizes the ranking procedure.
Moreover, it doesn't change the structure and training procedure of the network, and only increases negligible inference time.

\begingroup
\renewcommand{\arraystretch}{1}
  \begin{table*}[t]
    \begin{center}
      \caption{
        The ablation study of proposed methods.
        ``PairGAN" represents applying GAN to generate fake IR samples.
        ``Auxiliary'' represents jointly training the primary and auxiliary sets.
        ``Re-Ranking'' represents using the proposed temporal k-reciprocal re-ranking algorithm.
      }
      \label{table_ablation}
      \resizebox{0.95\textwidth}{!}{
        \begin{tabular}{cl|c|c|c|c|c|c|c|c|c|c|c|c|c}
          \toprule[1pt]
          & \multirow{2}{1.5cm}{\textbf{Method}} & \multirow{2}*{\textbf{PairGAN}} & \multirow{2}*{\textbf{Auxiliary}} 
          & \multirow{2}*{\textbf{Re-Ranking}}
          & \multicolumn{5}{c|}{\textbf{\textit{Visible to Infrared}}} & \multicolumn{5}{c}{\textbf{\textit{Infrared to Visible}}} \\
          \cline{6-15}
          ~ & ~ & ~ & ~ &
          & \textbf{R1} & \textbf{R5} & \textbf{R10} & \textbf{R20} & \textbf{mAP} 
          & \textbf{R1} & \textbf{R5} & \textbf{R10} & \textbf{R20} & \textbf{mAP} \\
          \hline
          & Baseline &      -     &      -     &      -     & 57.03 & 77.34 & 82.81 & 87.50 & 52.74 & 56.13 & 77.20 & 82.38 & 86.21 & 54.32 \\
          &          & \checkmark &            &            & 61.52 & 80.27 & 86.13 & 89.65 & 56.37 & 62.45 & 79.50 & 84.87 & 88.31 & 58.92 \\
          &          &            & \checkmark &            & 63.28 & 79.88 & 84.18 & 88.28 & 57.11 & 62.84 & 80.84 & 84.87 & 88.89 & 59.87 \\
          &          &            &            & \checkmark & 58.59 & 78.52 & 83.40 & 87.89 & 56.69 & 58.62 & 77.01 & 82.57 & 86.97 & 58.21 \\
          &          & \checkmark & \checkmark &            & 63.87 & 81.45 & 86.52 & 89.45 & 58.86 & 64.56 & 81.99 & 88.12 & 90.04 & 60.91 \\
          &          & \checkmark &            & \checkmark & 62.70 & 81.45 & 85.74 & 89.45 & 60.19 & 63.60 & 79.89 & 85.25 & 88.31 & 61.07 \\
          &          &            & \checkmark & \checkmark & 64.45 & 80.47 & 84.77 & 88.48 & 61.09 & 63.79 & 81.99 & 85.06 & 89.66 & 63.08 \\
          &          & \checkmark & \checkmark & \checkmark & \textbf{65.23} & \textbf{81.84} & \textbf{86.13} & \textbf{89.84} & \textbf{62.19} 
                                                            & \textbf{66.48} & \textbf{83.14} & \textbf{87.93} & \textbf{90.42} & \textbf{64.11} \\
          \bottomrule[1pt]
        \end{tabular}
      }
  \end{center}
\end{table*}
\endgroup

\begingroup
\renewcommand{\arraystretch}{1.3}
  \begin{table}[t]
    \begin{center}
      \caption{
        The impact of sampled sequence length in training and testing.
        The default setting is marked in \colorbox{lightgray}{gray}.
      }
      \label{table_length}
      \resizebox{0.4\textwidth}{!}{
        \begin{tabular}{lc|c|c|c|c}
          \toprule[1pt]
          & \multirow{2}*{\textbf{Sequence Length}} & \multicolumn{2}{c|}{\textbf{\textit{Visible to Infrared}}} & \multicolumn{2}{c}{\textbf{\textit{Infrared to Visible}}} \\
          \cline{3-6}
          &    & \textbf{Rank1} & \textbf{mAP} & \textbf{Rank1} & \textbf{mAP} \\
          \hline
          & 1  & 11.91 & 14.62 & 12.84 & 15.33 \\
          & 5  & 52.73 & 50.58 & 52.49 & 51.53 \\
          & \cellcolor{lightgray}10 & \cellcolor{lightgray}57.03 & \cellcolor{lightgray}52.74 & \cellcolor{lightgray}56.13 & \cellcolor{lightgray}54.32 \\
          & 15 & 58.98 & 54.79 & 58.81 & 56.42 \\
          & 20 & 57.62 & 55.02 & 60.15 & 58.13 \\
          \bottomrule[1pt]
        \end{tabular}
      }
    \end{center}
  \end{table}
\endgroup

\section{Experimental Results}

\subsection{Empirical Settings}
We conduct experiments on the constructed BUPTCampus dataset, 
in which 1,074 IDs are used for primary learning, 930 IDs for auxiliary learning, and 1,076 IDs for testing.
Rank1, Rank5, Rank10, Rank20 and mAP are adpoted as the evaluation metrics.

We implement our model by PyTorch \cite{paszke2019pytorch} and train it on two NVIDIA TESLA T4 GPUs.
ResNet-34 \cite{he2016deep} is utilized as the backbone because it performs better than ResNet-50 in our experiments.
Adam \cite{kingma2014adam} is utilized as the optimizer with the weight decay of 1e-5.
The learning rate is set as 2e-4 for initialization and updates with the cosine scheduler.
We set the maximum number of epochs to 100 and the batch size to 32 with $P=8$ identities and $K=4$ tracklets per identity.
The tracklet length is 10 and the image resolution is set to 256×128.
We use the random sampling strategy for training and the uniform sampling strategy for testing.
Random cropping and flipping are used for data augmentation.
The margin $m$ in triplet loss is set to 0.6, and the factor $\phi$ in Eq.\ref{alpha} is 3.
For the temporal k-reciprocal re-ranking algorithm, the intrinsic hyperparameters in the original k-reciprocal algorithm are set as $k1=5$, $k2=3$ and $\lambda_1=0.8$.
Then the newly added factor $\lambda_2$ is set to 0.1, and the number of groups is set as $L=2$.

\begingroup
\renewcommand{\arraystretch}{1.3}
  \begin{table*}[t]
    \begin{center}
      \caption{
        Comparison of our method with 9 state-of-the-art image-based and video-based methods on our BUPTCampus dataset.\\
        For image-based methods, we add a temporal average pooling operation to aggregate features from Seq\_Len=10 frames during testing for fair comparisons.
        For MITML$^\dagger$, we use SEQ\_LEN=6 instead of 10 because it yields better results in our experiments.\\
        ``w/o Aux'' represents that the auxiliary set is not used.
        CMC (\%) and mAP (\%) are reported.
      }
      \label{table_sota}
      \resizebox{0.95\textwidth}{!}{
        \begin{tabular}{cl|c|c|c|c|c|c|c|c|c|c|c|c|c}
          \toprule[1pt]
          & \multirow{2}{1cm}{\textbf{Type}} & \multirow{2}*{\textbf{Method}} & \multirow{2}*{\textbf{Reference}} 
          & \multirow{2}*{\textbf{Seq\_Len}}
          & \multicolumn{5}{c|}{\textbf{\textit{Visible to Infrared}}} & \multicolumn{5}{c}{\textbf{\textit{Infrared to Visible}}} \\
          \cline{6-15}
          ~ & ~ & ~ & ~ &
          & \textbf{R1} & \textbf{R5} & \textbf{R10} & \textbf{R20} & \textbf{mAP} 
          & \textbf{R1} & \textbf{R5} & \textbf{R10} & \textbf{R20} & \textbf{mAP} \\
          \hline
          & \multirow{16}*{Image}          & \multirow{2}*{AlignGAN\cite{wang2019rgb}}  & \multirow{2}*{ICCV2019}  & 1  & 26.11 & 44.81 & 52.59 & 59.81 & 28.01 & 20.52 & 37.87 & 46.83 & 53.73 & 23.25 \\
          &                                &                                            &                          & 10 & 35.37 & 53.89 & 61.30 & 68.70 & 35.13 & 27.99 & 49.07 & 57.65 & 66.60 & 30.32 \\
          \cline{3-15}
          &                                & \multirow{2}*{LbA\cite{park2021learning}}  & \multirow{2}*{ICCV2021}  & 1  & 25.93 & 48.15 & 57.78 & 66.67 & 28.98 & 25.00 & 43.66 & 51.87 & 61.75 & 27.07 \\
          &                                &                                            &                          & 10 & 39.07 & 58.70 & 66.48 & 75.37 & 37.06 & 32.09 & 54.85 & 65.11 & 72.57 & 32.93 \\ 
          \cline{3-15}
          &                                & \multirow{2}*{DDAG\cite{ye2020dynamic}}    & \multirow{2}*{ECCV2020}  & 1  & 33.15 & 56.85 & 64.63 & 71.11 & 33.85 & 29.29 & 50.93 & 59.51 & 68.47 & 32.03 \\
          &                                &                                            &                          & 10 & 46.30 & 68.15 & 74.44 & 81.30 & 43.05 & 40.44 & 40.86 & 61.38 & 69.78 & 78.54 \\
          \cline{3-15}
          &                                & \multirow{2}*{CAJ\cite{ye2021channel}}     & \multirow{2}*{ICCV2021}  & 1  & 36.67 & 54.63 & 63.70 & 70.19 & 35.26 & 29.85 & 53.73 & 61.75 & 69.03 & 32.29 \\
          &                                &                                            &                          & 10 & 45.00 & 70.00 & 77.04 & 83.33 & 43.61 & 40.49 & 66.79 & 73.32 & 81.16 & 41.46 \\
          \cline{3-15}
          &                                & \multirow{2}*{AGW\cite{ye2021deep}}        & \multirow{2}*{TPAMI2021} & 1  & 35.74 & 55.00 & 62.41 & 70.19 & 35.45 & 29.66 & 50.00 & 59.14 & 67.72 & 31.52 \\
          &                                &                                            &                          & 10 & 43.70 & 64.44 & 73.15 & 80.00 & 41.10 & 36.38 & 60.07 & 67.16 & 76.49 & 37.36 \\
          \cline{3-15}
          &                                & \multirow{2}*{MMN\cite{zhang2021towards}}  & \multirow{2}*{MM2021}    & 1  & 38.70 & 58.15 & 67.78 & 75.19 & 38.82 & 33.40 & 54.48 & 62.69 & 72.39 & 35.04 \\
          &                                &                                            &                          & 10 & 43.70 & 65.19 & 73.52 & 80.93 & 42.80 & 40.86 & 67.16 & 74.44 & 80.60 & 41.71 \\
          \cline{3-15}
          &                                & \multirow{2}*{DEEN\cite{zhang2023diverse}} & \multirow{2}*{CVPR2023}  & 1  & 40.19 & 65.56 & 73.52 & 79.44 & 40.70 & 35.63 & 59.70 & 68.28 & 76.49 & 38.67 \\
          &                                &                                            &                          & 10 & 53.70 & 74.81 & 80.74 & 87.59 & 50.43 & 49.81 & 71.64 & 80.97 & 85.82 & 48.59 \\
          \cline{3-15}
          &                                & \multirow{2}*{DART\cite{yang2022learning}} & \multirow{2}*{CVPR2022}  & 1  & 46.85 & 66.11 & 73.15 & 79.26 & 45.05 & 41.42 & 64.37 & 74.07 & 80.04 & 43.09 \\
          &                                &                                            &                          & 10 & 53.33 & 75.19 & 81.67 & 85.74 & 50.45 & 52.43 & 70.52 & 77.80 & 83.96 & 49.10 \\
          \hline
          & \multirow{2}*{Video}           & MITML$^\dagger$ \cite{lin2022learning}               & CVPR2022                 & 6  & 50.19 & 68.33 & 75.74 & 83.52 & 46.28 & 49.07 & 67.91 & 75.37 & 81.53 & 47.50 \\
          &                                & AuxNet (w/o aux)                           & ours                     & 10 & 62.70 & 81.45 & 85.74 & 89.45 & 60.19 & 63.60 & 79.89 & 85.25 & 88.31 & 61.07 \\
          &                                & \textbf{AuxNet}   &   \textbf{ours}   & \textbf{10} & \textbf{65.23} & \textbf{81.84} & \textbf{86.13} & \textbf{89.84} & \textbf{62.19} 
                                                                                                 & \textbf{66.48} & \textbf{83.14} & \textbf{87.93} & \textbf{90.42} & \textbf{64.11} \\
          \bottomrule[1pt]
        \end{tabular}
      }
    \end{center}
  \end{table*}
\endgroup

\begingroup
\renewcommand{\arraystretch}{1.3}
  \begin{table}[t]
    \begin{center}
      \caption{
        The ablation study of different strategies for the auxiliary factor $\alpha$.
        ``$E$'' is the normalized epoch index, from 0 to 1.
        The default setting is marked in \colorbox{lightgray}{gray}.
      }
      \label{table_auxiliary}
      \resizebox{0.45\textwidth}{!}{
        \begin{tabular}{lc|c|c|c|c|c}
          \toprule[1pt]
          & \small \multirow{2}*{\textbf{Strategy}} & \small \multirow{2}*{\textbf{Parameters}} & \multicolumn{4}{c}{\textbf{\textit{Infrared to Visible}}} \\
          \cline{4-7}
          &  &  & \textbf{mAP} & \textbf{Rank1} & \textbf{Rank5} & \textbf{Rank10} \\
          \hline
          & \normalsize \multirow{3}*{$fixed$}                                    & $\alpha=0.1$ & 55.05 & 57.85 & 76.05 & 81.42 \\
          &                                                                       & $\alpha=0.3$ & 57.82 & 59.77 & 80.08 & 84.10 \\
          &                                                                       & $\alpha=0.5$ & 55.26 & 56.13 & 77.39 & 82.18 \\
          \hline
          & \normalsize \multirow{3}*{$\alpha = {1\over 2} exp(-\tau E)$}         &   $\tau=1$   & 56.00 & 57.66 & 77.78 & 83.14 \\
          &                                                                       &   $\tau=3$   & 56.15 & 59.77 & 77.78 & 84.48 \\
          &                                                                       &   $\tau=5$   & 54.56 & 57.09 & 77.20 & 82.76 \\
          \hline
          & \normalsize \multirow{3}*{$\alpha = (cos(\pi E) + \phi) / 2(1+\phi)$} &   $\phi=1$   & 57.76 & 60.54 & 77.97 & 83.72 \\
          &                                                                       & \cellcolor{lightgray}$\phi=3$ & \cellcolor{lightgray}59.87 
                                                                                  & \cellcolor{lightgray}62.84 & \cellcolor{lightgray}80.84 & \cellcolor{lightgray}84.87 \\
          &                                                                       &   $\phi=5$   & 55.35 & 57.85 & 77.78 & 83.52 \\
          \bottomrule[1pt]
        \end{tabular}
      }
    \end{center}
  \end{table}
\endgroup

\begingroup
\renewcommand{\arraystretch}{1.3}
  \begin{table}[t]
    \begin{center}
      \caption{
        The comparison between the k-reciprocal algorithm and the proposed temporal k-reciprocal algorithm (marked in \colorbox{lightgray}{gray}).
        ``Time'' represents the inference time on all query and gallery samples. 
      }
      \label{table_reranking}
      \resizebox{0.45\textwidth}{!}{
        \begin{tabular}{lc|c|c|c|c|c|c}
          \toprule[1pt]
          & \multirow{2}*{\textbf{Method}} & \multirow{2}*{\textbf{re-ranking}} 
          & \multicolumn{4}{c|}{\textbf{\textit{Infrared to Visible}}} & \multirow{2}*{\textbf{Time(s)}} \\
          \cline{4-7}
          &                                            &              & \textbf{mAP} & \textbf{Rank1} & \textbf{Rank5} & \textbf{Rank10} \\
          \hline
          & \multirow{3}*{DDAG\cite{ye2020dynamic}}    &  \textbf{—}  & 40.44 & 40.86 & 61.38 & 69.78 & 349 \\
          &                                            & k-reciprocal & 42.03 & 41.23 & 61.19 & 68.10 & 350 \\
          &                                            & \cellcolor{lightgray}temporal & \cellcolor{lightgray}43.40 & \cellcolor{lightgray}42.35 
          & \cellcolor{lightgray}60.63 & \cellcolor{lightgray}67.54 & \cellcolor{lightgray}353 \\  
          \hline
          & \multirow{3}*{DART\cite{yang2022learning}} &  \textbf{—}  & 49.10 & 52.43 & 70.52 & 77.80 & 256 \\
          &                                            & k-reciprocal & 50.74 & 51.68 & 69.59 & 77.80 & 257 \\
          &                                            & \cellcolor{lightgray}temporal & \cellcolor{lightgray}52.01 & \cellcolor{lightgray}52.61 
          & \cellcolor{lightgray}68.47 & \cellcolor{lightgray}77.05 & \cellcolor{lightgray}260 \\  
          \hline
          & \multirow{3}*{MITML\cite{lin2022learning}} &  \textbf{—}  & 47.50 & 49.70 & 67.91 & 75.37 & 316 \\
          &                                            & k-reciprocal & 49.07 & 49.07 & 67.35 & 74.25 & 317 \\
          &                                            & \cellcolor{lightgray}temporal & \cellcolor{lightgray}51.30 & \cellcolor{lightgray}51.68 
          & \cellcolor{lightgray}69.78 & \cellcolor{lightgray}74.81 & \cellcolor{lightgray}321 \\  
          \hline
          & \multirow{3}*{Baseline (ours)}             &  \textbf{—}  & 54.32 & 56.13 & 77.20 & 82.38 & 70 \\
          &                                            & k-reciprocal & 56.08 & 56.51 & 76.82 & 81.99 & 71 \\
          &                                            & \cellcolor{lightgray}temporal & \cellcolor{lightgray}58.21 & \cellcolor{lightgray}58.62 
          & \cellcolor{lightgray}77.01 & \cellcolor{lightgray}82.57 & \cellcolor{lightgray}74 \\
          \bottomrule[1pt]
        \end{tabular}
      }
    \end{center}
  \end{table}
\endgroup

\begin{table*}[t]
  \begin{center}
    \caption{
      The impact of parameter selection of the proposed temporal k-reciprocal algorithm for the ``Infrared to Visible'' setting.
      The default parameters are $k1$=5, $k2=3$, $\lambda_1$=0.8, $\lambda_2$=0.1 and $L$=2.
    }
    \label{table_parameter}
      
    \begin{minipage}{0.33\textwidth}
    \centering
      \resizebox{0.95\textwidth}{!}{
        \begin{tabular}{lc|c|c|c|c}
          \toprule[1pt]
          & $\pmb{L}$ & \textbf{mAP} & \textbf{Rank1} & \textbf{Rank5} & \textbf{Rank10} \\
          \hline
          & 1  & 56.08 & 56.51 & 76.82 & 81.99 \\
          & \cellcolor{lightgray}2 & \cellcolor{lightgray}58.21 & \cellcolor{lightgray}58.62 & \cellcolor{lightgray}77.01 & \cellcolor{lightgray}82.57 \\
          & 4 & 58.73 & 59.00 & 78.54 & 83.52 \\
          & 8 & 59.13 & 59.58 & 79.50 & 83.52 \\
          \bottomrule[1pt]
        \end{tabular}
        \label{a}
      }
    \end{minipage}
    \begin{minipage}{0.33\textwidth}
      \resizebox{0.95\textwidth}{!}{
        \begin{tabular}{lc|c|c|c|c}
          \toprule[1pt]
          & $\pmb{k1}$ & \textbf{mAP} & \textbf{Rank1} & \textbf{Rank5} & \textbf{Rank10} \\
          \hline
          & 3 & 58.06 & 58.43 & 77.59 & 82.95 \\
          & 4 & 58.02 & 58.62 & 77.39 & 82.38 \\
          & \cellcolor{lightgray}5 & \cellcolor{lightgray}58.21 & \cellcolor{lightgray}58.62 & \cellcolor{lightgray}77.01 & \cellcolor{lightgray}82.57 \\
          & 6 & 57.88 & 57.85 & 77.01 & 82.76 \\
          & 7 & 57.79 & 57.66 & 77.97 & 82.95 \\
          \bottomrule[1pt]
        \end{tabular}
      }
    \end{minipage}
    \begin{minipage}{0.33\textwidth}
      \resizebox{0.95\textwidth}{!}{
        \begin{tabular}{lc|c|c|c|c}
          \toprule[1pt]
          & $\pmb{k2}$ & \textbf{mAP} & \textbf{Rank1} & \textbf{Rank5} & \textbf{Rank10} \\
          \hline
          & 1 & 56.86 & 58.05 & 77.78 & 83.33 \\
          & 2 & 58.17 & 57.85 & 77.78 & 83.14 \\
          & \cellcolor{lightgray}3 & \cellcolor{lightgray}58.21 & \cellcolor{lightgray}58.62 & \cellcolor{lightgray}77.01 & \cellcolor{lightgray}82.57 \\
          & 4 & 57.24 & 57.28 & 77.97 & 82.57 \\
          & 5 & 56.52 & 56.32 & 77.01 & 82.38 \\
          \bottomrule[1pt]
        \end{tabular}
      }
    \end{minipage}

    \vspace{5mm}
    
    \begin{minipage}{0.33\textwidth}
      \resizebox{0.95\textwidth}{!}{
        \begin{tabular}{lc|c|c|c|c}
          \toprule[1pt]
          & $\pmb{\lambda_1}$ & \textbf{mAP} & \textbf{Rank1} & \textbf{Rank5} & \textbf{Rank10} \\
          \hline
          & 0.6 & 57.79 & 57.28 & 75.48 & 81.42 \\
          & 0.7 & 58.00 & 57.66 & 76.44 & 81.99 \\
          & \cellcolor{lightgray}0.8 & \cellcolor{lightgray}58.21 & \cellcolor{lightgray}58.62 & \cellcolor{lightgray}77.01 & \cellcolor{lightgray}82.57 \\
          & 0.9 & 57.96 & 59.00 & 77.20 & 83.33 \\
          & 1.0 & 57.19 & 58.62 & 77.78 & 82.95 \\
          \bottomrule[1pt]
        \end{tabular}
      }
    \end{minipage}  
    \begin{minipage}{0.33\textwidth}
      \resizebox{0.95\textwidth}{!}{
        \begin{tabular}{lc|c|c|c|c}
          \toprule[1pt]
          & $\pmb{\lambda_2}$ & \textbf{mAP} & \textbf{Rank1} & \textbf{Rank5} & \textbf{Rank10} \\
          \hline
          & 0    & 56.08 & 56.51 & 76.82 & 81.99 \\
          & \cellcolor{lightgray}0.1 & \cellcolor{lightgray}58.21 & \cellcolor{lightgray}58.62 & \cellcolor{lightgray}77.01 & \cellcolor{lightgray}82.57 \\
          & 0.2 & 58.68 & 59.00 & 77.39 & 82.95 \\
          & 0.3 & 58.52 & 58.81 & 77.78 & 82.57 \\
          & 0.4 & 58.45 & 57.66 & 77.01 & 81.80 \\
          \bottomrule[1pt]
        \end{tabular}
      }
    \end{minipage}

  \end{center}
\end{table*}

\subsection{Ablation Study}
\noindent{\textbf{Main Results.}}
Tab.\ref{table_ablation} summarizes the overall ablation results of the main components of our methods, i.e., PairGAN, auxiliary learning and temporal re-ranking.
We can draw the following observations:
\begin{itemize}
  \item The introduced PairGAN module can bring 4\% to 6\% Rank1 improvement and approximate 4\% mAP improvement over baseline,
  which validates it can help the modality-invariant learning.
  \item The designed auxiliary learning method improves Rank1 by 6\% and mAP by 5\% respectively,
  which indicates the effectiveness of exploiting the single-camera samples in training.
  \item The temporal re-ranking algorithm can improve the mAP by 4\%.
  That means it can effectively revise the ranking order with mined gallery-to-gallery similarities and temporal correlations.
  \item These above components complement each other. By integrating them all, 
  our final methods improve the performance of baseline by approximately 10\% Rank1 and 10\% mAP. 
\end{itemize}

\noindent{\textbf{Sequence Length.}}
Compared with image-based ReID, video-based ReID can expolit rich temporal information,
which helps eliminate effects of noises and distractors.
To verify the necessity of video data, we study the influence of sequence length in Tab.\ref{table_length},
and we have the following conclusions:
\begin{enumerate}
  \item A terrible result is obtained in case of setting sequence length to 1, which means the task degrades to image-based ReID. 
  That indicates the image-based methods cannot work well in our benchmark.
  \item As the sequence length varies from 1 to 15, the performance is gradually improved,
  which demonstrates the effectiveness of temporal information.
  \item When the sequence length is larger than 15, the Rank1 performance degrades under the ``visible-to-infrared'' mode,
  which shows that excessive length will bring more redundant information.   
\end{enumerate}
To achieve a balance between accuracy and efficiency, we set the sequence length to 10 by default.

\noindent{\textbf{Auxiliary Learning.}}
We compare different designs of the curriculum factor $\alpha$ in our auxiliary learning method in Tab.\ref{table_auxiliary}.
Three types of design are listed, i.e., fixed value, exponential decline and cosine decline.
For a fixed value, $\alpha=0.3$ achieves a trade-off between the primary task and auxiliary task.
As to the exponential decline design, it achieves a similar performance to the fixed value.
Differently, a cosine decline strategy performs much better, with an improvements of 2\% mAP and 3\% Rank1 over them.

\noindent{\textbf{Temporal Re-ranking.}}
The comparisions between the original k-reciprocal re-ranking algorithm and our proposed temporal re-ranking algorithm are listed in Tab.\ref{table_reranking}.
Four different models are selected as the baseline methods, 
i.e., DDAG\cite{ye2020dynamic}, DART\cite{yang2022learning}, MITML\cite{lin2022learning} and our constructed baseline network.
It is shown that the original k-reciprocal algorithm can consistently improve the mAP metrics, but sometimes decreases Rank1.
Differently, our temporal re-rakning algorithm can further increase both mAP and Rank1 by a remarkable margin.
Moreover, it only increases negligible inference time.
We further study the effects of parameter selection in Tab.\ref{table_parameter}.
As reported, the performance is robust to the varying values of $k1, k2, \lambda_1, \lambda_2$ and $L$.

\definecolor{myBlue}{RGB}{127, 127, 255}
\definecolor{myRed2}{RGB}{255, 127, 127}
\begin{figure*}[t]
  \centering
  \includegraphics*[width=1\textwidth]{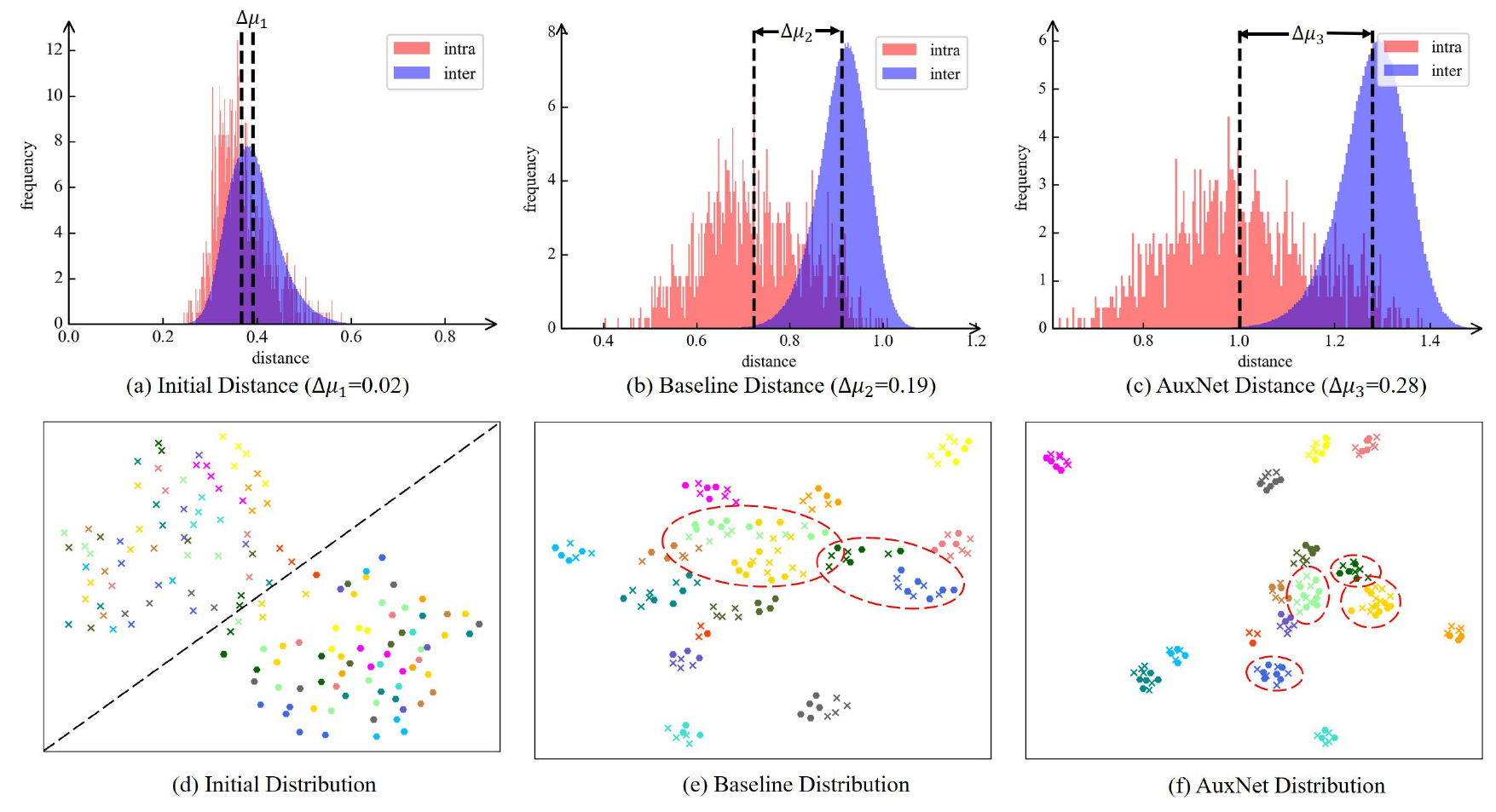}
  \caption{
    The distribution of distances and features of the initial model, baseline and our AuxNet on the test set of BUPTCampus.
    \textbf{(a-c)} The distributions of the intra-class distances (in \colorbox[RGB]{255, 127, 127}{red}) 
    and inter-class distances (in \colorbox[RGB]{127, 127, 255}{blue}). 
    $\Delta \mu$ is the difference between the means of the two types of distances.
    Larger $\Delta \mu$ represents stronger discriminative ability.
    \textbf{(d-e)} Visualization of the corresponding feature space by UMAP\cite{mcinnes2018umap}.
    The samples with the same color are from the same ID.
    The ``cross" and ``dot" markers denote the RGB and IR modalities, respectively.
  }
  \label{fig_dist}
\end{figure*}

\subsection{Main Results.}
For comprehensive comparisions, we reproduce 9 other state-of-the-art methods on BUPTCampus as shown in Tab.\ref{table_sota}.
For image-based methods,
i.e., AlignGAN\cite{wang2019rgb}, LbA\cite{park2021learning}, DDAG\cite{ye2020dynamic}, CAJ\cite{ye2021channel},
AGW\cite{ye2021deep}, MMN\cite{zhang2021towards}, DEEN\cite{zhang2023diverse}, DART\cite{yang2022learning},
the default setting is the sequence length of 1 during inference.
For a fair comparision, we also report their results with the length of 10,
which is implemented by adding a temporal average pooling operation during inference.
For the video-based method, MITML\cite{lin2022learning}, we report the results with the length of 6,
as it performs better than the length of 10 in our implements.
The results indicate that our solution, dubbed \textbf{AuxNet}, shows substantial superiority to existing state-of-the-art methods.

Furthermore, we conduct experiments on the HITSZ-VCM dataset and the results are shown in Tab.\ref{table_vcm}.
Please note that the proposed ``PairGAN'' and ``auxiliary learning'' methods cannot be utilized here.
Therefore, only the results of baseline (based on ResNet50) and temporal re-ranking are givem.
The parameters of temporal re-ranking are set to $k1$=20, $k2$=6, $\lambda_1$=0.3 and $\lambda_2$=0.4.

\begin{figure}[t]
  \centering
  \includegraphics[width=0.4\textwidth]{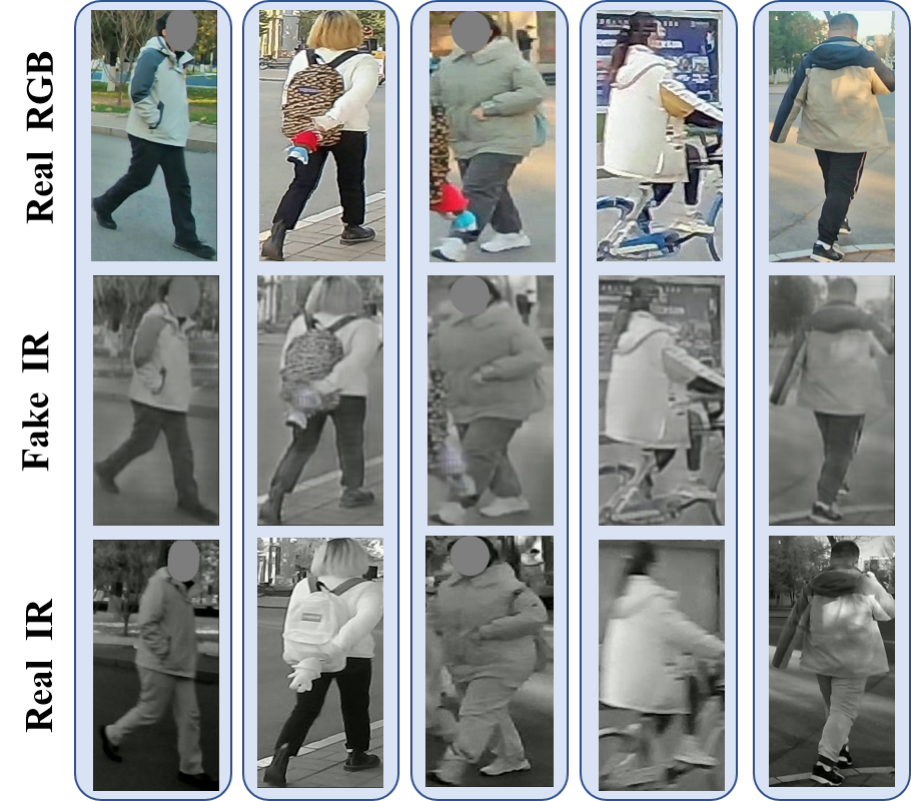}
  \caption{
    Visualization of sampled real RGB, generated fake IR and real IR images.
    Same ID for the same column.
  }
  \label{fig_fakeir}
\end{figure}

\begin{figure}[t]
  \centering
  \includegraphics[width=0.4\textwidth]{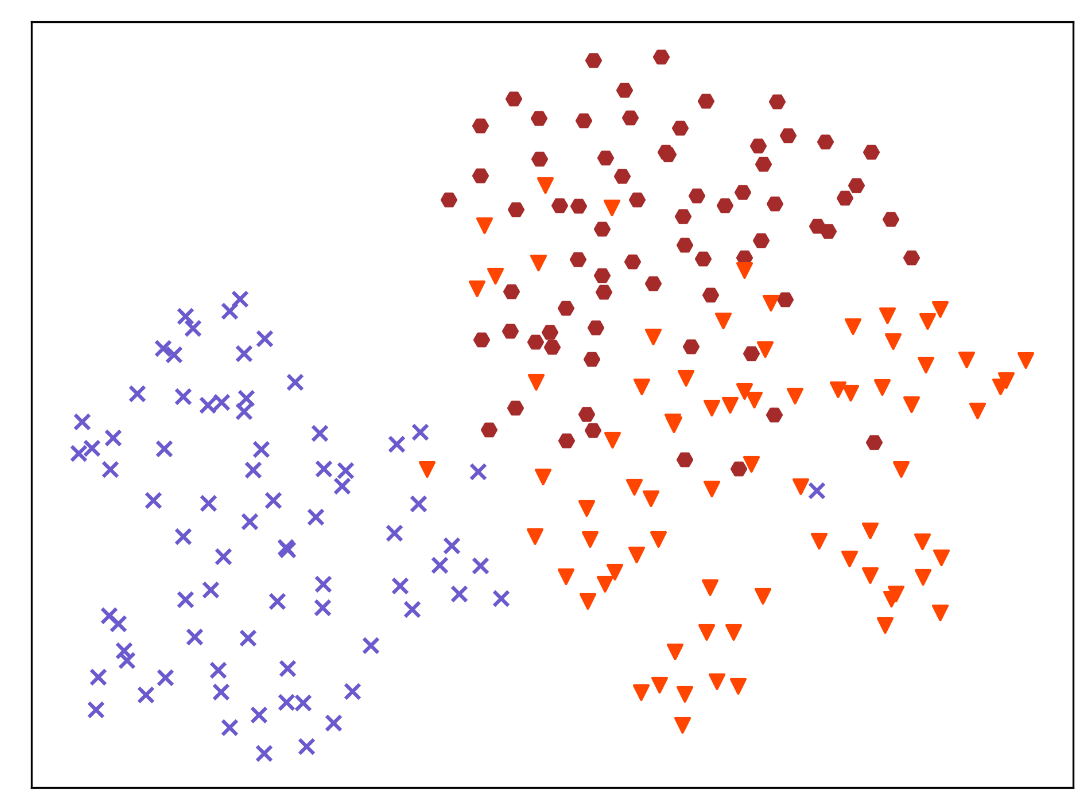}
  \caption{
    The distribution of real RGB (blue ``cross''), fake IR (red ``triangle'') and real IR (brown ``dot'') samples visualized by UMAP \cite{mcinnes2018umap}.
  }
  \label{fig_dist_2}
\end{figure}

\definecolor{myGreen}{RGB}{226,240,217}
\definecolor{myRed}{RGB}{241,219,217}
\begin{figure}[t]
  \centering
  \includegraphics*[width=0.4\textwidth]{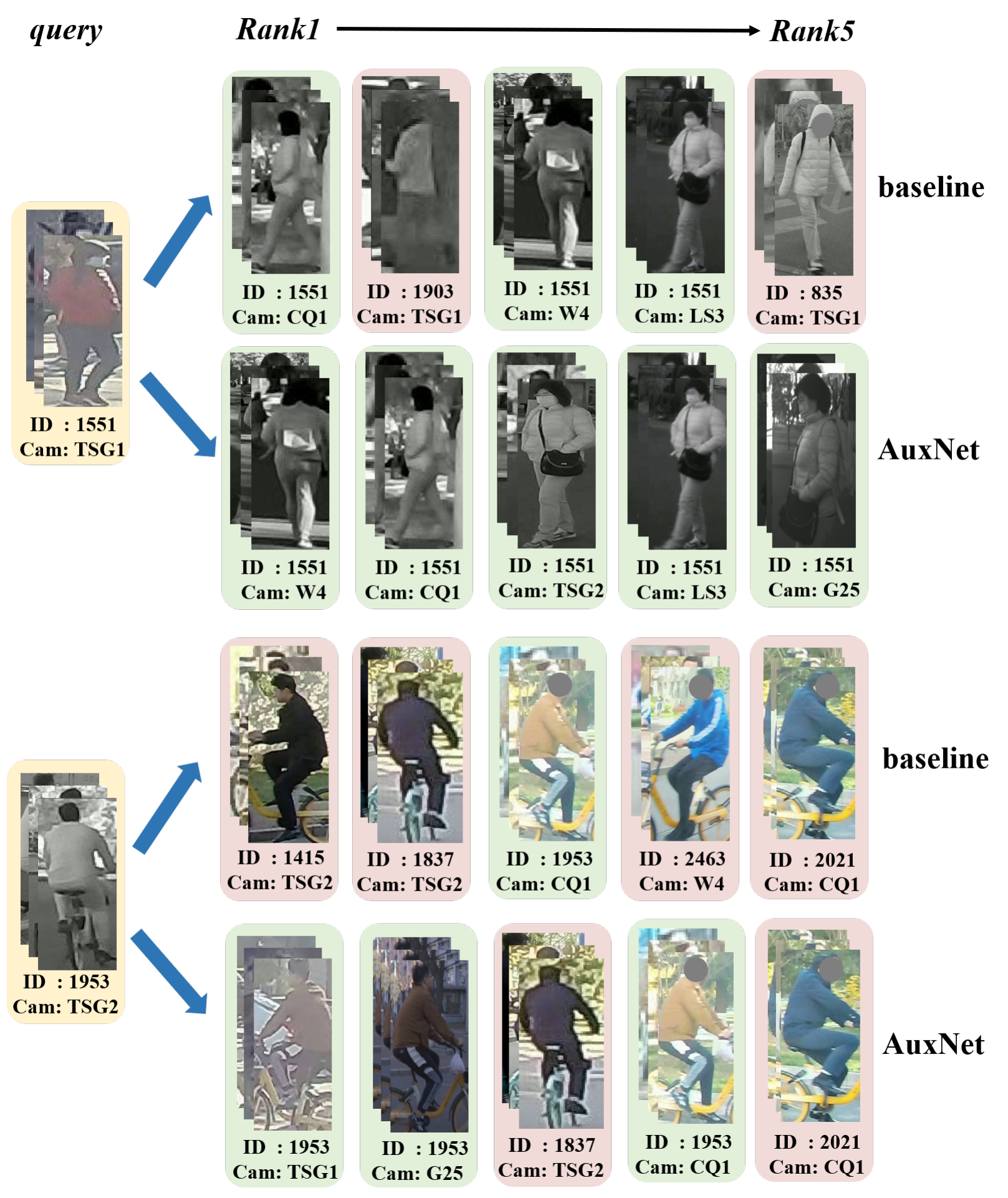}
  \caption{
    Sampled retrieval results of our baseline and proposed AuxNet on the BUPTCampus.
    For one query, the top-5 ranking samples are visualized.
    The right samples are marked in \colorbox[RGB]{226,240,217}{green}, and the false samples are in \colorbox[RGB]{241,219,217}{red}. 
  }
  \label{fig_list}
\end{figure}

\begingroup
\renewcommand{\arraystretch}{1.3}
  \begin{table}[t]
    \begin{center}
      \caption{
        Comparison of our method with other state-of-the-art methods on the HITSZ-VCM dataset.
        Please note that our proposed ``PairGAN'' and  ``auxiliary learning'' cannot be utilized here.
      }
      \label{table_vcm}
      \resizebox{0.45\textwidth}{!}{
        \begin{tabular}{lc|c|c|c|c}
          \toprule[1pt]
          & \small \multirow{2}*{\textbf{Method}} & \multicolumn{2}{c|}{\textbf{\textit{Visible to Infrared}}} & \multicolumn{2}{c}{\textbf{\textit{Infrared to Visible}}} \\
          \cline{3-6}
          &  & \textbf{mAP} & \textbf{Rank1} & \textbf{mAP} & \textbf{Rank1} \\
          \hline
          & LbA (ICCV 2021) \cite{park2021learning} & 32.38 & 49.30 & 30.69 & 46.38 \\
          & MPANet (CVPR 2021) \cite{wu2021discover} & 37.80 & 50.32 & 35.26 & 46.51 \\
          & DDAG (ECCV 2020) \cite{ye2020dynamic} & 41.50 & 59.03 & 39.26 & 54.62 \\
          & VSD (CVPR 2021) \cite{tian2021farewell} & 43.45 & 57.52 & 41.18 & 54.53 \\
          & CAJ (ICCV 2021) \cite{ye2021channel} & 42.81 & 60.13 & 41.49 & 56.59 \\
          & MITML (CVPR 2022) \cite{lin2022learning} & 47.69 & 64.54 & 45.31 & 63.74 \\
          \hline
          & \textbf{Baseline (ours)} & \textbf{36.90} & \textbf{52.34} & \textbf{36.26} & \textbf{49.32} \\
          & \textbf{+ temporal re-ranking} & \textbf{48.70} & \textbf{54.58} & \textbf{45.99} & \textbf{51.05} \\
          \bottomrule[1pt]
        \end{tabular}
      }
    \end{center}
  \end{table}
\endgroup

\subsection{Visualization.}
\noindent{\textbf{Feature Distribution.}}
In order to further analyze the performance difference between baseline and our AuxNet, 
we compute the feature distance distribution and feature space distribution.
Fig.\ref{fig_dist}(a) shows the initial distance distributions of the intra-class and inter-class pairs,
in which features are extracted by the network without training on BUPTCampus.
Fig.\ref{fig_dist}(b) and (c) visualize the distance distributions obtained by baseline and AuxNet respectively.
It is shown that out method can further separate the intra-class and inter-class distances compared with baseline,
which has a larger difference $\Delta \mu$ of the means of the two distributions.
In Fig.\ref{fig_dist}(d-f), we plot the feature embeddings in the 2D feature space for visualization using UMAP \cite{mcinnes2018umap}.
The results show that our AuxNet can better distinguish difficult negative samples (marked by red dotted ellipses),
and greatly narrows the gap between the two modality samples with the same identity.

\noindent{\textbf{PairGAN.}}
Fig.\ref{fig_fakeir} visualizes the sampled real RGB, generated fake IR by PairGAN and corresponding real IR images.
The fake IR images preserve the detailed texture cues from the RGB modality, and have the similar color style with the IR modality.
Therefore, it can help alleviate the differences between the two modalities and learn modality-invariant embeddings.
We further visualize the distribution of samples of these three modalities in Fig.\ref{fig_dist_2}.
It can be observed that compared to real RGB samples, the fake IR samples have a closer distribution to real IR samples.

\noindent{\textbf{Ranking List.}}
Fig.\ref{fig_list} compares the ranking lists of baseline and our AuxNet in both ``visible-to-infrared'' and ``infrared-to-visible'' settings.
In the first case, given an RGB query sample (ID 1551), the baseline method retrieves two wrong samples in the top-5 ranking list.
Instead, our AuxNet successfully revises them. 
Specifically, the positive sample under camera ``G25'' has huge modality and view differences from the query sample, and is correctly retrieved.
In the second case with an IR query sample (ID 1953), our AuxNet achieves the right top-1 and top-2 matching results.

\section{Conclusions}

In this paper, we contribute a new benchmark for video-based visible-infrared person re-identification, named BUPTCampus,
which is the first dataset with RGB/IR tracklet pairs and auxiliary samples.
It consists of 3,080 identities, 1,869,066 images and 16,826 tracklets.
Furthermore, we construct a two-stream network as baseline and present the PairGAN module to help modality-invariant learning.
To exploit the auxiliary samples, we design to train primary samples and auxiliary samples jointly with a curriculum factor.
Finally, we propose a novel temporal k-reciprocal algorithm to re-rank the retrieval results with fine-grained temporal correlation cues.
We demonstrate the effectiveness of our method by comparing it with 9 state-of-the-art works.
We hope the contributed dataset and methods can help to narrow the gap between academic works and realistic applications.

\section*{Acknowledgments}
This work is supported by Chinese National Natural Science Foundation under Grants (62076033, U1931202)
and BUPT Excellent Ph.D. Students Foundation (CX2022145).

\bibliographystyle{splncs04}
\bibliography{egbib}
\end{document}